\documentclass[11pt]{article}

\usepackage{acl}

\usepackage{times}
\usepackage{latexsym}
\usepackage[T1]{fontenc}
\usepackage[utf8]{inputenc}
\usepackage{microtype}
\usepackage{inconsolata}
\usepackage{graphicx}
\usepackage{booktabs}
\usepackage{amsmath}
\usepackage{amssymb}
\usepackage{tikz}
\usetikzlibrary{positioning,arrows.meta,calc,fit,backgrounds}
\definecolor{predc}{RGB}{37,84,158}   
\definecolor{obsc}{RGB}{198,106,20}   
\definecolor{gray20}{RGB}{225,227,232}

\title{Team eye-be-am at SHROOM-Visions: Two-Token Features and
Small--Large Ensembles for VLM Hallucination Detection}

\author{Eli Schwartz \\
  IBM Research \\
  \texttt{eliyahu.schwartz@ibm.com} \\}

\begin{document}
\maketitle

\begin{abstract}
We present our system for the SHROOM-Visions 2026 shared task on
character-level VLM hallucination detection. A small ($4$B-parameter)
VLM is fine-tuned as a per-token classifier reading a two-token feature
from its own hidden states, and is ensembled with a $\sim$400B zero-shot
VLM judge at prediction time. Both components see off-the-shelf OCR of
any visible in-image text. We use synthetic hallucination data generated
by the large model as a source of ensemble diversity, and use
validation to select feature layer, training data and OCR grounding.
Our official entry reaches mean Cor $0.487$ / Cor-lbl $0.387$ on the
hidden test set, placing $6$th/$28$ (EN), $6$th/$21$ (FR), $8$th/$21$
(IT) and $7$th/$22$ (ZH) on the task's primary Cor-lbl metric.
\end{abstract}

\section{Introduction}
\label{sec:intro}

The SHROOM-Visions 2026 shared task~\cite{shroom2026overview} asks systems
to identify, at character granularity, which spans of a VLM's textual output
are hallucinated with respect to a supplied image and prompt, and to
classify each span into one of five categories (\emph{invention},
\emph{mischaracterization}, \emph{OCR}, \emph{miscounting}, \emph{other}).
Scoring is by two per-language (EN, FR, IT, ZH) Spearman correlations: an
unlabelled $\rho$ (``Cor'') and a label-aware $\rho_{\text{lbl}}$
(``Cor-lbl'').

Two solution families present themselves. A very large instruction-tuned
VLM can be prompted in zero shot to name the hallucinated spans directly;
this generalises across languages but is noisy per character. Fine-tuning a
purpose-built detector on the shared-task data is more reliable per token,
but fine-tuning at the $\sim$400B-parameter scale against 15k labelled
examples is not compute-efficient. We instead fine-tune a VLM two orders of
magnitude smaller than the zero-shot judge and ensemble the two at
prediction time.

The paper is organised around what we could measure on validation data.
First, we introduce a two-token classifier head that concatenates the
hidden states of the current and previous tokens at a single language-tower
layer (\S\ref{sec:head}) and sweep the choice of layer to identify a
mid-network state as the strongest feature source (\S\ref{sec:layers}).
Second, we use synthetic hallucination data generated by the large model
as a source of \emph{ensemble diversity}, measuring the token-level
disagreement between small and large models that accompanies it
(\S\ref{sec:synth}). Third, we ground both components with off-the-shelf
OCR of any in-image text (\S\ref{sec:ocr}) and A/B-test the lift.
\S\ref{sec:results} reports the val-based ablation ladder that stacks
these three ingredients, and \S\ref{sec:submission} reports how the
resulting system performed on the hidden test set.

\section{Task and Models}
\label{sec:models}

Each datapoint is $(x, p, y)$: an image $x$, a prompt $p$, and a
VLM-generated text $y$. For each character position $i$ in $y$ the system
predicts a probability $\hat{p}_i \in [0,1]$ that the character is inside
a hallucinated span, and a category
$\hat{c}_i \in \{\text{inv}, \text{misch}, \text{OCR}, \text{count},
\text{other}, \text{none}\}$; character-level probability series are
scored against per-character averaged annotator gold using the official
2026 scorer. The SHEEP dataset behind the shared task~\cite{mickus2026sheep}
provides human-written span annotations across four languages (EN, FR,
IT, ZH) over outputs from five VLMs.

\paragraph{Small model.} A 4B-parameter Qwen3.5-VL instruction-tuned
checkpoint. Visual encoder and multimodal connector are frozen;
LoRA adapters~\cite{hu2022lora} ($r{=}16$, $\alpha{=}32$) are attached to the MLP submodules
of all 32 language-model layers.

\paragraph{Large model.} Qwen3.5-VL-397B (A17B MoE, FP8 serving) used
zero-shot via a JSON-schema prompt that asks for an explicit list of
hallucinated span strings with categories. The same model, invoked
offline, also synthesises training data for the small model
(\S\ref{sec:synth}).

\paragraph{OCR extractor.} Both models additionally see an off-the-shelf
OCR extraction of visible in-image text, injected verbatim into the user
prompt (\S\ref{sec:ocr}).

\section{Small-Model Detector}
\label{sec:small}

\subsection{A two-token feature}
\label{sec:head}

\begin{figure}[t]
\centering
\resizebox{\columnwidth}{!}{%
\begin{tikzpicture}[
  font=\small,
  layer/.style={draw=black!45, fill=white, rounded corners=1.5pt,
                minimum width=2.0cm, minimum height=0.55cm, inner sep=1pt},
  hl/.style={draw=#1, line width=0.9pt, fill=#1!12},
  tok/.style={layer, fill=gray20, draw=black!35},
  vd/.style={fill=white, inner sep=2pt, minimum width=0.5cm},
  op/.style={draw=black!55, fill=black!4, rounded corners=2.5pt,
             minimum width=2.4cm, minimum height=0.62cm, inner sep=3pt},
  flow/.style={-{Stealth[length=2.6mm]}, line width=0.9pt},
]
\def\xa{0}      
\def\xb{3.6}    

\node[tok]             (a0)  at (\xa,0)     {$x_{t-1}$};
\node[layer]           (a1)  at (\xa,0.80)  {\footnotesize L1};
\node[vd]              (ad1) at (\xa,1.42)  {$\vdots$};
\node[layer, hl=predc] (a16) at (\xa,2.04)  {\footnotesize L16};
\node[vd]              (ad2) at (\xa,2.66)  {$\vdots$};
\node[layer]           (a31) at (\xa,3.28)  {\footnotesize L32};

\node[tok]             (b0)  at (\xb,0)     {$x_{t}$};
\node[layer]           (b1)  at (\xb,0.80)  {\footnotesize L1};
\node[vd]              (bd1) at (\xb,1.42)  {$\vdots$};
\node[layer, hl=obsc]  (b16) at (\xb,2.04)  {\footnotesize L16};
\node[vd]              (bd2) at (\xb,2.66)  {$\vdots$};
\node[layer]           (b31) at (\xb,3.28)  {\footnotesize L32};

\begin{scope}[on background layer]
  \foreach \x in {\xa,\xb}{
    \draw[-{Stealth[length=2.4mm,width=2.2mm]}, black!35, line width=1.1pt]
          (\x,-0.42) -- (\x,3.02);
  }
\end{scope}

\node[align=center, font=\small] at (\xa,4.30) {token $t\!-\!1$};
\node[align=center, font=\small] at (\xb,4.30) {token $t$};

\node[op] (cat) at (7.7,3.05) {concat};
\node[op] (mlp) at (7.7,2.00) {MLP $\;2d\!\to\!d\!\to\!6$};
\node[op, minimum width=2.9cm] (sm) at (7.7,0.95) {softmax};

\draw[flow, predc] (a16.west) -- (-1.5,2.04) -- (-1.5,3.85)
                   -- (7.7,3.85) -- (cat.north);
\node[predc, anchor=south, font=\small] at (5.6,3.89) {$h^{(16)}_{t-1}$};

\draw[flow, obsc] (b16.east) -- (6.15,2.04) -- (6.15,3.05) -- (cat.west);
\node[obsc, anchor=south, font=\small] at (5.35,2.10) {$h^{(16)}_{t}$};

\draw[flow] (cat) -- (mlp);
\draw[flow] (mlp) -- (sm);

\node[font=\scriptsize, anchor=north] at (7.7,0.52)
      {6-way hallucination label per token};

\end{tikzpicture}%
}
\caption{Two-token classifier head: for token $t$ we concatenate
\textcolor{predc}{$h_{t-1}^{(16)}$} with \textcolor{obsc}{$h_t^{(16)}$}
and pass the pair to a shallow MLP.}
\label{fig:head}
\end{figure}
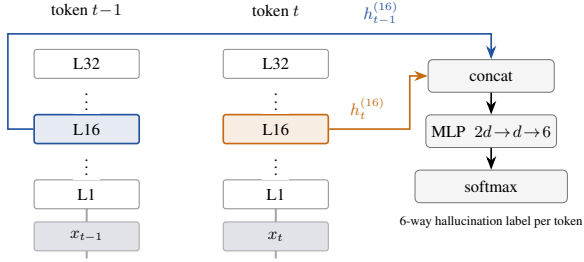

For every generated token $t$ in the output $y$ we form a per-token
feature by concatenating hidden states of the current and previous tokens
at the same language-tower layer:
\begin{equation}
z_t = \bigl[\, h_t^{(L)} \,\Vert\, h_{t-1}^{(L)} \,\bigr],
\label{eq:head}
\end{equation}
and a two-layer MLP over $z_t$ predicts a $6$-way categorical distribution
(five hallucination categories plus \emph{none}); Figure~\ref{fig:head}
shows the extraction schematically. The current token contributes an
image-conditioned reading of what actually appeared; the previous token
contributes local context that is inexpensive to concatenate and
consistently helpful in our controls. Which language-tower layer to read
is the more consequential choice, and we set it by sweep: $L = 16$ of
the $33$ hidden states the backbone exposes ($h^{(0)}$ is the embedding
output, $h^{(32)}$ the final block; \S\ref{sec:layers}).

Because $y$ comes from a different, undisclosed VLM and we consume it
teacher-forced, the feature probes hidden states of a \emph{reviewing}
model rather than the generating one, unlike prior hidden-state
probes~\cite{azaria2023saplma,chen2024inside}.

\subsection{Which layer to read}
\label{sec:layers}

We select $L$ empirically: with architecture, data, schedule and seed
fixed we sweep $L$ over seven values from the embedding output to the
penultimate block on val (Table~\ref{tab:layer-sweep}). Mean Cor rises
by $+0.039$ from $L{=}0$ to a broad plateau centred on the middle of
the tower and then flattens, and we adopt the plateau midpoint
$L{=}16$; a follow-up run tying both slots to $L{=}16$ (rather than
only $h_t$) adds a further $+0.017$ (App.~\ref{app:layer-placement}).
EN-only controls confirm that the previous-token slot is doing work:
replacing $h_{t-1}$ by a second copy of $h_t$ costs at most $0.008$
Cor, while permuting the second slot along time---preserving its
marginal statistics but destroying position alignment---costs $0.049$.

\begin{table}[h]
\centering
\small
\setlength{\tabcolsep}{4pt}
\begin{tabular}{ccccccc}
\toprule
$L$ & EN & FR & IT & ZH & Cor & lbl \\
\midrule
0  & 0.319 & 0.285 & 0.312 & 0.416 & 0.333 & 0.304 \\
6  & 0.330 & 0.297 & 0.329 & 0.421 & 0.344 & 0.312 \\
13 & 0.340 & 0.312 & 0.354 & 0.430 & 0.359 & 0.322 \\
\textbf{16} & 0.348 & 0.330 & 0.365 & 0.446 & \textbf{0.372} & 0.328 \\
19 & 0.351 & 0.329 & 0.358 & 0.452 & 0.372 & 0.331 \\
26 & 0.347 & 0.310 & 0.351 & 0.451 & 0.365 & 0.328 \\
31 & 0.352 & 0.317 & 0.365 & 0.448 & 0.371 & \textbf{0.333} \\
\bottomrule
\end{tabular}
\caption{Val Cor per current-token feature layer $L$ (head-only, frozen
base, 1 epoch, four languages joint). Layer indices are the backbone's
\texttt{hidden\_states} indices ($0$ = embedding output, $32$ = final
block). The mid-network state $L{=}16$ is the mean-Cor best and
$L{=}19$ is essentially tied; the curve is broad but distinctly favours
the middle of the tower over the edges (embedding-output $+0.039$ Cor).}
\label{tab:layer-sweep}
\end{table}

\subsection{Training and inference}
\label{sec:training}

The classifier head and LoRA adapter are trained jointly with plain 6-way
cross-entropy over the token grid, warm-started from a single-epoch
head-only checkpoint, on all four languages jointly
(App.~\ref{app:hyper}). At inference the small model is run
teacher-forced over the provided output text; per-token distributions are
projected to characters, and we report $\hat{p}_i = 1 - P(\text{none})$
with $\hat{c}_i = \arg\max_{c \neq \text{none}} P(c)$.

\subsection{OCR grounding}
\label{sec:ocr}

The backbone's visual encoder reads clean prominent text reliably but is
noisy on stylised captions, small print and occluded text---the material
behind the task's \emph{OCR} category. We factor that failure mode out by
running an off-the-shelf OCR engine~\cite{du2020paddleocr} per language and
injecting its output verbatim as a labelled \texttt{EXTRACTED IMAGE TEXT:}
block in both models' user message (omitted when empty); weights are
unchanged and the prompts present the string as a signal, not ground truth.
On val this lifts the large judge's mean $\rho$ from $0.264$ to $0.303$ in
all four languages (strongest on IT and ZH, where in-image reading is
weakest); a matched EN A/B on the small classifier lifts $\rho$ from $0.331$
to $0.342$, concentrated on rows with non-empty extractions ($+0.021$ on
$n{=}149$ of $401$).

\section{Synthetic Data for Ensemble Diversity}
\label{sec:synth}

The 15k shared-task training examples do not, by themselves, ensure that the
small model produces signal \emph{complementary} to the large judge. For
squared-error ensembles the ambiguity
decomposition~\cite{krogh1994ensembles} makes the point precise: gain over
the average constituent equals the constituents' disagreement, so making the
small model more accurate is not the same as making the ensemble better. Our
metric is a rank correlation, so we use this for motivation and for the proxy
it suggests rather than as an identity we verify. We therefore generate extra
training data with the large model, aiming to shift the small model's error
distribution away from the large model's own.

\paragraph{Pipeline.} A single multimodal prompt to the 397B model, given an
image and its caption, returns JSON with three fields: a clean faithful
description; a re-written version perturbing one or more spans to introduce
a hallucination of a specified category; and the list of changed phrases
with categories. Spans come from a character-diff between the two
descriptions, falling back to approximate string matching when the diff
count disagrees with the declared changes ($\approx 41\%$ of rows). This
yields $\approx 67{,}500$ labelled rows across four languages, a
$\approx 68\%$ usable rate on 24k input captions per language.

Using the same large model to both generate training data and serve as
the judge at inference does not, we argue, undermine ensemble diversity.
The two roles invoke the model under different prompts and against
different inputs: the generator is asked to \emph{produce} an image
caption plus a targeted perturbation from an image and short caption
seed, while the judge is asked to \emph{annotate} a fixed VLM output
against the same image with per-span categories. Training the small
model on the generator's output therefore teaches it what
generator-style hallucinations look like, not the judge's per-token
scoring function. In practice this leaves room for the two systems to
disagree, and the disagreement measurement below shows that room is
non-trivially used.

\paragraph{Does it change the small model?}
We mix synthetic with real rows at a $0.5$ ratio during small-model training.
Whether that diversifies the small model is empirical: for every token in the
four-language val split ($140{,}565$ tokens, $1{,}533$ samples) we compare
the two models' argmax hallucination flags (any non-\emph{none} class).
Disagreement rises from $28.9\%$ (real-only) to $30.1\%$ with
synth---a modest $+1.2$~pp. Argmax-over-six-classes and mean L1 distance
between the 6-way distributions are essentially unchanged, so the effect is
on \emph{which} tokens each model flags, not on category re-shuffling within
already-flagged tokens. Whether the additional diversity translates into
ensemble gain is measured in \S\ref{sec:results}.

\section{Results}
\label{sec:results}

At prediction time we ensemble the small model's predicted probability
$\hat{p}^{S}_i$ and the large model's zero-shot probability
$\hat{p}^{L}_i$ at each character position $i$ by a token-level convex
combination
\begin{equation}
\hat{p}_i \;=\; \alpha\,\hat{p}^{S}_i + (1-\alpha)\,\hat{p}^{L}_i,
\label{eq:ensemble}
\end{equation}
with $\alpha{=}0.8$ fixed; the predicted category comes from whichever
model contributes the majority of the mass at position $i$. On val the
sweep over $\alpha$ (Table~\ref{tab:alpha}) is flat across
$\alpha \in [0.8, 1.0]$ and drops sharply below $\alpha \approx 0.5$
where the large judge's coarse span-level probabilities dominate the
character series. On the hidden test set the fixed global
$\alpha{=}0.8$ also outperformed per-language val-tuned weights.
\label{sec:ensemble}

\begin{table}[h]
\centering
\small
\setlength{\tabcolsep}{6pt}
\begin{tabular}{ccc}
\toprule
$\alpha$ & mean Cor & mean $\rho_{\text{lbl}}$ \\
\midrule
0.0 & 0.253 & 0.182 \\
0.4 & 0.260 & 0.185 \\
0.5 & 0.330 & 0.239 \\
0.6 & 0.428 & 0.339 \\
0.7 & 0.457 & 0.377 \\
\textbf{0.8} & \textbf{0.462} & 0.387 \\
0.9 & 0.462 & \textbf{0.391} \\
1.0 & 0.457 & 0.387 \\
\bottomrule
\end{tabular}
\caption{Val mean Cor and mean $\rho_{\text{lbl}}$ vs.\ ensemble weight
$\alpha$ (Eq.~\ref{eq:ensemble}; $\alpha{=}1.0$ is head-only,
$\alpha{=}0.0$ is large-model-only). Selected rows shown; the plateau
on $[0.8, 1.0]$ is flat.}
\label{tab:alpha}
\end{table}

\begin{table}[t]
\centering
\footnotesize
\setlength{\tabcolsep}{2pt}
\begin{tabular}{@{}lcccccc@{}}
\toprule
& EN & FR & IT & ZH & Cor & lbl \\
\midrule
Small, fine-tuned    & 0.401 & 0.417 & 0.445 & 0.543 & 0.451 & 0.379 \\
${}+{}$synth         & 0.418 & 0.432 & 0.438 & 0.538 & 0.457 & 0.388 \\
${}+{}$OCR           & 0.425 & 0.453 & 0.452 & 0.538 & \textbf{0.467} & \textbf{0.398} \\
Ensemble             & 0.430 & 0.461 & 0.454 & 0.534 & 0.470 & 0.394 \\
\midrule
Large, 0-shot, OCR   & 0.264 & 0.223 & 0.276 & 0.386 & 0.287 & 0.287 \\
\bottomrule
\end{tabular}
\caption{Val Cor per language and mean Cor / Cor-lbl (``lbl'') for the
small detector at $L{=}16$. Rows in the top block share architecture,
schedule and seed and differ only in the data and prompt they see; each
row adds to the one above. Data and OCR grounding are monotone in both
metrics; the ensemble with the large judge is essentially flat on val
(see text). The zero-shot large judge is listed at the bottom for
reference and is fused into the row above it at $\alpha{=}0.8$
(\S\ref{sec:ensemble}).}
\label{tab:ablation}
\end{table}

\paragraph{Val ablation.} Table~\ref{tab:ablation} stacks the three
ingredients of \S\ref{sec:head}--\S\ref{sec:ocr}: synthetic data and OCR
grounding each add $\approx {+}0.010$ mean Cor-lbl on top of the small
model at $L{=}16$, and the best small detector reaches $0.467$ mean Cor
/ $0.398$ Cor-lbl. The diversity measurement of \S\ref{sec:synth}
follows the same rungs: token-level hallucination-flag disagreement
between the small model and the OCR-grounded large judge rises from
$28.9\%$ (real-only) to $30.1\%$ (real+synth). On val, however, the
ensemble with the large judge is essentially flat: both scoring metrics
award a
vacuous match when gold and prediction are simultaneously empty, and the
OCR-grounded large judge lands \emph{at} that empty-labels floor
($0.287$ mean; Table~\ref{tab:ablation}, bottom row). By the ambiguity
decomposition~\cite{krogh1994ensembles} an ensemble gains only when both
constituents contribute complementary signal, and on val the judge does
not.

\subsection{Submission results}
\label{sec:submission}

On the hidden test set the picture is different: the large judge scores
$0.355$ mean Cor, well above its val floor, and the two constituents
become useful ensemble partners. The most likely explanation for the
val--test gap is annotation quality: the val split is a hash-based
holdout of the released training data and inherits its label noise,
whereas the hidden test set is separately curated by the task
organisers, and the large zero-shot judge---already producing
label-quality output---appears to be penalised on val for disagreements
that annotator noise would forgive on test.

\begin{table}[t]
\centering
\footnotesize
\setlength{\tabcolsep}{2.5pt}
\begin{tabular}{@{}lcccccc@{}}
\toprule
& EN & FR & IT & ZH & Cor & lbl \\
\midrule
Large, 0-shot     & 0.434 & 0.313 & 0.331 & 0.342 & 0.355 & 0.265 \\
Small, fine-tuned & 0.418 & 0.469 & 0.472 & 0.457 & 0.454 & 0.356 \\
Ensemble          & 0.460 & 0.481 & 0.485 & 0.512 & \textbf{0.485} & \textbf{0.382} \\
\midrule
Official entry    & 0.466 & 0.484 & 0.485 & 0.512 & \textbf{0.487} & \textbf{0.387} \\
Rank (Cor-lbl)    & 6/28 & 6/21 & 8/21 & 7/22 & --- & --- \\
\bottomrule
\end{tabular}
\caption{Hidden-test Cor per language and means for the three main test
entries and our final official submission. Ensemble fuses the row above
it with the large judge at $\alpha{=}0.8$; the small model there is our
best small-only upload on Cor-lbl (real+synth). The official-entry row
is a per-language Cor-lbl best of our uploads: an OCR-grounded ensemble
on FR and the ensemble of the row above on EN, IT and ZH. Rank is by
Cor-lbl.}
\label{tab:submission}
\end{table}

The large judge is the stronger system on EN and the weaker one on the
other three languages by $0.12$ to $0.14$; fusing them adds $+0.031$
mean Cor over the small model and $+0.130$ over the judge, and wins on
three languages of four. Our official entry, a per-language Cor-lbl best
across uploaded configurations, reaches mean Cor $\mathbf{0.487}$ /
Cor-lbl $\mathbf{0.387}$ (Table~\ref{tab:submission}), placing $6$th,
$6$th, $8$th and $7$th of $28$, $21$, $21$ and $22$ teams. Cor rank
equals or beats Cor-lbl rank in every language, so span detection is our
stronger half and categorisation the weaker.

\section{Related Work}\label{sec:related}

The previous SHROOM edition~\cite{shroom2025} scored text-only hallucination
with the same per-character metrics; the 2026 edition adds visual failure
modes---OCR errors and miscounting---and character-granularity spans remain
unusual against the whole-response or object-level judgements common in
VLM-hallucination benchmarking~\cite{li2023pope,sun2024mmhalbench}. Probing
internal states for factuality is established---SAPLMA~\cite{azaria2023saplma}
classifies truthfulness from activations,
INSIDE~\cite{chen2024inside} exploits internal consistency---and our head
sits in that lineage. Related mid-network decoding
work~\cite{chuang2024dola} contrasts earlier and later layer distributions
to reduce hallucination, whereas our head simply reads a mid-network state
as a feature. The ensemble reading borrows the ambiguity
decomposition~\cite{krogh1994ensembles}, here for teacher-forced token
classifiers, where diversity is directly measurable.

\section{Conclusion}

A small fine-tuned VLM with a two-token classifier head reading a
mid-network hidden state gives a competitive per-character hallucination
detector on the SHROOM-Visions 2026 shared task. Layer choice is the
largest lever ($+0.039$ val Cor from the embedding output to the mid
layer); OCR grounding and synthetic training data each add a further
$+0.010$ mean Cor on val. On the hidden test set the small detector
ensembled with a $\sim$400B zero-shot VLM judge lifts mean Cor by
$+0.132$ over the judge alone and $+0.033$ over the small detector alone,
placing the system in the top third of every language.\mbox{}\label{endmain}

\section*{Limitations}\label{sec:limitations}

Every configuration was trained once, with seed $42$; per-language swings
from synth training reach $0.023$ Cor, larger than several of the deltas
we discuss, so single-language comparisons should be read as descriptive.
The placement gain of App.~\ref{app:layer-placement} is measured head-only
and is an upper bound on what layer choice contributes to the full
system---LoRA and more data absorb most of it. The diversity account of
\S\ref{sec:synth} predicts the test ensemble gain but not the val
outcome, because on val the large judge sits at the empty-labels floor
and cannot combine usefully; on test it is $0.07$ Cor above that floor
and the ensemble gains. Our submitted system also differs from the
system this paper describes: it was trained at a cross-layer placement
rather than the same-layer $L{=}16$ we adopt, folded the val holdout
into training, and its official-entry row is a per-language Cor-lbl best
across configurations rather than one runnable system. App.~\ref{app:extended-limitations}
gives the full accounting; App.~\ref{app:negative-results} lists
test-time augmentation strategies that did not transfer into ensemble
gain. Finally, our study uses a single model family on both sides of the
ensemble (Qwen3.5-VL 4B and 397B) and a single OCR extractor
(PaddleOCR), and the synthetic set is badly category-mismatched to real
data---the most likely reason categorisation is our weakest metric.

\bibliography{shroom2026}

\clearpage
\appendix

\section{Prompts used}
\label{app:prompts}

All three prompts are quoted verbatim from the shipped code
(\texttt{shroom/judge/prompts.py} and
\texttt{scripts/vllm\_397b\_combined\_synth.py}).

\paragraph{Large-model zero-shot judge (system + user template).}
\begin{small}
\begin{verbatim}
[SYSTEM]
You are a strict hallucination detector for
vision-language model outputs. You will see an
image, the prompt that was given to a VLM, and
the VLM's output. Your job is to identify spans
of the output text that are NOT supported by
the image -- i.e., hallucinations.

Categories:
A. Invention  -- entities, objects, or events
                 not present in the image.
B. Mischaracterization -- incorrect description
                          of something visible.
C. OCR Problem -- misreading of text visible in
                  the image.
D. Miscounting -- incorrect reporting of
                  quantities of visible items.
E. Other -- does not fit A-D.

Return STRICT JSON only -- no prose, no markdown
fences. The JSON is a single object with one key,
"spans", whose value is an array. Each span is an
object with three keys:
  - "span": the literal substring from the OUTPUT
            TEXT that is hallucinated. Must be
            quoted exactly as it appears,
            character-for-character.
  - "confidence": a float in [0, 1]. Higher = more
                  confident this span is
                  hallucinated.
  - "category": one of "A", "B", "C", "D", "E".

If nothing is hallucinated, return {"spans": []}.

Example output:
{"spans": [{"span": "three cats",
            "confidence": 0.85,
            "category": "D"}]}

[USER]
PROMPT TO THE VLM:
{prompt}

VLM OUTPUT TEXT:
{output_text}

Identify hallucinated spans in the OUTPUT TEXT.
Respond with JSON only.
\end{verbatim}
\end{small}

\paragraph{Synthetic-data generator (large 397B model, per-language).}
The generator is called once per row on the CC6M image--caption corpus.
The system prompt is language-specialised via an in-line placeholder;
the shown template is for English.
\begin{small}
\begin{verbatim}
[SYSTEM]
You are a data augmentation tool. You see an
image and a short English caption.
Produce ONE JSON object with three keys:

  "clean":  A natural longer description of the
            image in English, 30-500 characters.
            Ground the description in what is
            actually visible. Do not add facts
            not shown.
  "edited": The clean description with EXACTLY
            these edits applied in order:
              [edit spec 1]
              [edit spec 2]
              ...
            Length 30-500 chars. Do not reorder
            sentences unless the edit requires it.
  "changes": A JSON array with one entry per
             edit, in the same order.
             Each entry is
             {"category": "<one of invention|
                          mischaracterization|
                          OCR|miscounting|other>",
              "phrase":   "<exact substring from
                          'edited' that changed>"}

Return STRICT JSON only. No prose, no markdown
fences.

[USER]
CAPTION: {caption}
[image attached]
\end{verbatim}
\end{small}
The edit-spec placeholders are filled with a fixed edit-instruction
per category (e.g. \emph{invention}: ``Add a new plausible-in-context
object, entity, or attribute NOT mentioned in the source caption'').

\paragraph{Classifier system prompt (small-model training and inference).}
The system message prepended to every training and inference example:
\begin{small}
\begin{verbatim}
You are a hallucination detector for vision-
language model outputs. Given an image and a
VLM response, judge which characters in the
response are hallucinated and, when they are,
into which category they fall.
\end{verbatim}
\end{small}

\paragraph{OCR-grounded user prompt (both small classifier and large
zero-shot judge).}
The user turn seen by both models is the concatenation of the VLM
prompt, an optional \texttt{EXTRACTED IMAGE TEXT:} block (omitted when
the OCR extractor returns the empty string), and the VLM output text:
\begin{small}
\begin{verbatim}
PROMPT TO THE VLM:
{prompt}

EXTRACTED IMAGE TEXT:
{ocr_text}

VLM OUTPUT TEXT:
{output_text}
\end{verbatim}
\end{small}
Detected OCR line-strings are joined with `` | ''; when the extractor
returns the empty string the entire block is omitted and the prompt
reduces to the OCR-less form. For the large judge, the classifier
system prompt above is followed by the additional instruction that
\emph{extracted image text may be noisy on stylised or rotated text and
should be treated as a signal, not ground truth}.

\section{Hyperparameters and Compute}
\label{app:hyper}

\paragraph{Small-model training.}
\begin{itemize}
\setlength{\itemsep}{0pt}
\item Base: Qwen3.5-VL 4B, visual encoder and multimodal connector frozen;
      only the language-tower LoRA adapters + classifier head are trained.
\item LoRA: $r{=}16$, $\alpha{=}32$, applied to the three MLP
      projections \texttt{\{gate,up,down\}\_proj} of every
      language-model layer (96 modules total). Attention projections
      are left untouched.
\item Head: 2-layer MLP over the concatenated feature
      $z_t = [h_t^{(16)}, h_{t-1}^{(32)}]$ (Eq.~\ref{eq:head}), output
      dimension 6.
\item Optimiser: AdamW, learning rate $2\!\times\!10^{-4}$, cosine
      schedule over total steps, no warmup, weight decay $0.0$,
      $(\beta_1,\beta_2) = (0.9, 0.999)$.
\item Batch: per-device batch size $2$, gradient accumulation $8$;
      effective batch $16$ on single-GPU, $32$ under 2-GPU DDP.
\item Loss: unweighted 6-way categorical cross-entropy over the token
      grid. Preliminary runs with flat inverse-frequency class weights
      collapsed category prediction onto the majority class; unweighted
      CE was more stable.
\item Precision: bfloat16 forward + backward, SDPA attention.
\item Warm-start: the head is initialised from a 1-epoch head-only
      training run (base frozen, no LoRA); the LoRA adapter is fresh.
\item Data: shared-task train split (four languages jointly), plus a
      $0.5$ mix of the synthetic training set (\S\ref{app:data}), plus
      OCR-augmented prompts on every row (\S\ref{sec:ocr}).
      Images resized so the long side is $448$ pixels.
\item Val: hash-based $10\%$ held out for model selection and diversity
      measurement (\S\ref{sec:synth}); once the
      configuration is fixed the val slice is folded back into training
      for the submission run, since with OCR grounding the small model
      has a new signal to learn from an additional $10\%$ of labelled
      data.
\item Epochs: 2. Seed 42.
\end{itemize}

\paragraph{Compute.}
Each small-model training run consumes $\approx 4$ H100-GPU-hours (single
H100, $\sim$2.5\,h wall for 2 epochs with synth mix=0.5). Test-time
inference for the small model takes $\approx 1.5$ H100-GPU-hours
end-to-end across four languages. Large-model zero-shot judge predictions
on the test set are obtained via vLLM on 8 H100s at $\sim$1.7\,s/sample
with \texttt{enable\_thinking=False}, totalling $\approx 3$
H100-node-hours ($\approx 24$ H100-GPU-hours) across the four languages.

The dominant cost is synthetic-data generation, and we state it plainly
because it is easy to under-report. Generating the synthetic set required
one 397B multimodal call per input caption over $24$k captions in each of
four languages, i.e.\ $\approx 96$k calls at $\sim$1.7\,s/sample on the
same 8-H100 configuration: $\approx 45$ H100-node-hours, or
$\approx 363$ H100-GPU-hours. That is roughly $90\times$ the cost of
training the small model and about $85\%$ of the total compute behind our
official entry. The small model is therefore cheap to \emph{train} and
cheap to \emph{serve}, but the pipeline that produced it is not cheap; we
make no claim that the system as a whole is compute-efficient relative to
alternatives that skip synthetic generation. A practitioner reproducing
only the real-data ensemble (mean Cor $0.467$) pays the $\approx 4$
GPU-hour training cost and none of the $363$.

\section{Data Statistics}
\label{app:data}

\begin{table}[h]
\centering
\small
\setlength{\tabcolsep}{4pt}
\begin{tabular}{lrrrr}
\toprule
Split & EN & FR & IT & ZH \\
\midrule
Train (real, 90\%)  & 3{,}398 & 3{,}386 & 3{,}384 & 3{,}401 \\
Val (real, 10\%)    & 401     & 381     & 362     & 389 \\
Test (hidden)       & 1{,}201 & 1{,}233 & 1{,}254 & 1{,}210 \\
\midrule
Synth (mix=0.5)     & 16{,}459 & 16{,}600 & 16{,}415 & 18{,}041 \\
\bottomrule
\end{tabular}
\caption{Sample counts per split. Val is a deterministic hash-based
10\% held-out slice of the released training data (ids with a fixed
SHA-256-derived hash below the cutoff).}
\label{tab:data-counts}
\end{table}

The synthetic set is produced by the pipeline of \S\ref{sec:synth} run
once per language on 24k input captions from CC6M, yielding a
$\approx 68\%$ usable-row rate. The fuzzy-alignment path fires on
$\approx 41\%$ of rows across all four languages.

\section{Same-Layer vs.\ Cross-Layer Follow-Up}
\label{app:layer-placement}

Table~\ref{tab:layer-sweep} pins the previous-token slot at a fixed
reference layer and sweeps only the current-token layer $L$; the head
at $L{=}16$ there is $[h_t^{(16)} \Vert h_{t-1}^{(L_b)}]$. To choose
between tying the two feature slots and keeping them cross-layer we
re-run the head-only training at $L{=}16$ under both placements,
holding architecture, data, schedule and seed fixed.

\begin{table}[h]
\centering
\small
\setlength{\tabcolsep}{3pt}
\begin{tabular}{lcccccc}
\toprule
placement & EN & FR & IT & ZH & Cor & lbl \\
\midrule
cross-layer & 0.348 & 0.330 & 0.365 & 0.446 & 0.372 & 0.328 \\
same-layer  & \textbf{0.358} & \textbf{0.364} & \textbf{0.377} & \textbf{0.459} & \textbf{0.389} & \textbf{0.333} \\
\bottomrule
\end{tabular}
\caption{Val Cor for the two head-only placements at $L{=}16$: cross-layer
draws $h_{t-1}$ from a fixed reference layer, same-layer draws both slots
from $L{=}16$. Same-layer gains $+0.017$ mean Cor in all four languages.
This is a head-only delta and an upper bound on what placement contributes
to the full system (\S\ref{sec:limitations}).}
\label{tab:layer-placement}
\end{table}

\section{Extended Limitations}
\label{app:extended-limitations}

\paragraph{Placement is head-only.} The $+0.017$ Cor same-layer advantage
of Table~\ref{tab:layer-placement} is measured head-only on a frozen
base. Re-running the val ladder with LoRA enabled, the advantage of
same-layer $[16\Vert16]$ over cross-layer $[16\Vert32]$ shrinks from
$+0.005$ mean Cor on the real-only rung to $0.000$ once synthetic data
and OCR grounding are added, both placements reaching $0.467$ mean Cor
at the top rung ($+0.003$ Cor-lbl for $[16\Vert16]$). A trainable
adapter and more data evidently recover most of what the better
placement supplies.

\paragraph{Diversity story and val--test asymmetry.} The disagreement
measurement of \S\ref{sec:synth} is a val-side observation, but the
ensemble Cor lift it predicts does not appear on val. We attribute this
to the val--test asymmetry of the large judge: on val it sits at the
empty-labels floor, so fusion finds little to combine even when
disagreement is present. On test the judge is $0.07$ above that floor,
and the ensemble gains $+0.031$ mean Cor over the small model of
Table~\ref{tab:submission}. The synth-driven part of that gain is
however not test-only clean: the labelled channel of the small model
also improves on test with synth ($0.345 \rightarrow 0.356$ Cor-lbl), so
we do not claim a purely diversity-driven effect.

\paragraph{OCR arm is confounded on test.} The OCR arm of our submission
folded the val holdout into training in the same run, so its test delta
is the joint effect of OCR grounding and $10\%$ more labelled data, not
of OCR alone. The clean OCR measurements are the val A/Bs in
\S\ref{sec:ocr}.

\paragraph{Composite submission.} The Official-entry row of
Table~\ref{tab:submission} is a per-language Cor-lbl best of our uploads
rather than one runnable system: an OCR-grounded ensemble on FR and the
ensemble of the row above on EN, IT and ZH. Because the task selects on
Cor-lbl, its Cor is not our maximum (ZH: $0.512$ against $0.518$ for the
OCR ensemble). The submitted small detector was trained at the
cross-layer $[16\Vert32]$ placement rather than the same-layer $L{=}16$
selected in \S\ref{sec:layers}. The small model of
Table~\ref{tab:submission} is our best small-only upload on Cor-lbl but
not on Cor---an earlier configuration on real-only data scores $0.457$
mean test Cor against this row's $0.454$ while losing on Cor-lbl
($0.349$ against $0.356$), so the honest reading of the ensemble's Cor
gain over the best small model we ever uploaded is $+0.028$ rather than
$+0.031$.

\paragraph{Reproducibility of the EN test row.} The EN figures for our
final ensemble differ between the values returned to us at upload time
($0.460$/$0.347$) and the final leaderboard values in
Table~\ref{tab:submission} ($0.466$/$0.357$); the IT and ZH figures for
the same submitted file agree to four decimals across the two views. We
cannot account for the EN difference.

\paragraph{Category-distribution mismatch.} The synthetic set is badly
distribution-mismatched at the category level:
\emph{mischaracterization} accounts for $38$--$44\%$ of real spans but
at most $6\%$ of synthetic ones (and $0\%$ in IT and ZH), while
\emph{other} is $3$--$6\%$ of real spans and $\approx 28\%$ of synthetic
ones; \emph{miscounting} and \emph{OCR} are likewise under-generated.
This is the most likely explanation for why categorisation is our
weakest metric.

\paragraph{Compute and ensemble weight.} The ensemble weight $\alpha$
was fixed rather than learned; a learned per-token gate is a natural
extension we did not pursue. Our system is also small only in its
trainable part: the large judge is required at test time, and
synthetic-data generation consumes roughly $90\times$ the large-model
compute that small-model training consumes (App.~\ref{app:hyper}), so we
make no end-to-end efficiency claim.

\section{Negative Results}
\label{app:negative-results}

We record three probes that did not translate into ensemble gain and are
kept here as pointers for others considering similar directions.

\paragraph{Image multi-scale TTA.}
Because the output text is fixed at test time we cannot apply standard
sampling-based test-time augmentation. On the image side we retried the
small model at three long-side resizes ($384$, $448$, $512$), averaged
the resulting per-token 6-way distributions, and re-scored. On EN val
this changes the argmax on $0.5\%$ of tokens---roughly two orders of
magnitude less than the $28.9\%$ small--large inter-model disagreement
that drives the ensemble---and mean Cor is unchanged to four decimals.
The image encoder produces essentially the same reading at plausible
scales; multi-scale averaging is not a useful diversity source in this
pipeline.

\paragraph{MC-dropout on LoRA adapters.}
Enabling dropout at inference within the LoRA adapters and averaging $K$
forward passes gives a bounded form of head-side TTA. With
\texttt{lora\_dropout}$=0.1$ and $K{=}10$ we see $0.17\%$ argmax
disagreement across passes on EN val; sweeping the inference dropout
rate to $\{0.2, 0.3, 0.5\}$ raises this to $0.34\%$, $0.52\%$ and
$0.76\%$ respectively---monotone in $p$ but still an order of magnitude
below the inter-model diversity level. The contribution of the LoRA
adapter to the final prediction is bounded, so dropping fractions of it
produces less variance than swapping the entire small model for the
large judge.

\paragraph{Verdict-position log-probabilities as a signal.}
Before adopting the classifier-head architecture, we explored a
JSON-completion setup in which the small model was fine-tuned to emit a
per-sentence \emph{verdict} token (hallucinated / not) and we recovered a
per-sentence $P(u)$ from the top-$K$ output log-probabilities at the
verdict position. On EN val the resulting soft probabilities were
worse-calibrated to gold than the head architecture we eventually
adopted: fused with a strong span-selection channel, verdict-logprob
$P(u)$ \emph{hurt} the $\rho$ metric by $0.09$ mean. The verdict
position's uncertainty was not gold-correlated: models that had been
trained to emit ``no hallucination'' with high frequency did so with
high confidence regardless of whether the sentence was in fact clean.
We abandoned this pathway.

\end{document}